\documentclass[fullpaper,cameraready]{nldl}

\usepackage[utf8]{inputenc}
\usepackage{url}
\usepackage{authblk}
\usepackage[square,numbers]{natbib} 
\usepackage{algorithm}
\usepackage{algorithmic}

\usepackage{times}  
\usepackage{helvet}  
\usepackage{courier}  
\usepackage{graphicx} 
\usepackage{natbib}  
\usepackage[font=scriptsize]{caption} 

\usepackage{amsmath}
\title{Expert Q-learning: Deep Reinforcement Learning with Coarse State Values from Offline Expert Examples}
\author[1]{Li Meng}
\author[3,4,5]{Anis Yazidi}
\author[2,3]{Morten Goodwin}
\author[1]{Paal Engelstad}
\affil[1]{University of Oslo}
\affil[2]{Centre for Artificial Intelligence Research, University of Agder}
\affil[3]{Oslo Metropolitan University}
\affil[4]{Norwegian University of Science and Technology}
\affil[5]{Oslo University Hospital}

\date{\vspace{-5ex}}

\begin{document}
\nldlmaketitle

\begin{abstract}  
In this article, we propose a novel algorithm for deep reinforcement learning named Expert Q-learning. Expert Q-learning is inspired by Dueling Q-learning and aims at incorporating semi-supervised learning into reinforcement learning through splitting Q-values into state values and action advantages. We require that an offline expert assesses the value of a state in a coarse manner using three discrete values. An expert network is designed in addition to the Q-network, which updates each time following the regular offline minibatch update whenever the expert example buffer is not empty. Using the board game Othello, we compare our algorithm with the baseline Q-learning algorithm, which is a combination of Double Q-learning and Dueling Q-learning. Our results show that Expert Q-learning is indeed useful and more resistant to the overestimation bias. The baseline Q-learning algorithm exhibits unstable and suboptimal behavior in non-deterministic settings, whereas Expert Q-learning demonstrates more robust performance with higher scores, illustrating that our algorithm is indeed suitable to integrate state values from expert examples into Q-learning.
\end{abstract}

\section{Introduction}
Reinforcement learning (RL) is one of the machine learning (ML) research areas that studies approaches to endowing agents with intelligence using trial and error. RL, unlike supervised learning, resorts to  a reward function and has no need of labeled data. An agent only requires (partial) information about the current state observation of the environment to choose actions so as to maximize the cumulative reward. Nevertheless, finding an optimal solution is not always possible and RL algorithms seek for a balance between the amount of exploitation and exploration \cite{kaelbling1996reinforcement}. Namely, an agent chooses whether to maximize its return based on past experience, or to explore non-visited states which can have potentially higher returns. A large amount of exploitation can lead to sub-optimal solutions, yet excessive exploration leads to slow convergence of the trajectories.

Deep reinforcement learning (DRL) combines RL with deep learning (DL) techniques. Neural networks (NNs), especially deep convolutional neural networks (CNNs), can be trained together with the search routines of RL \cite{krizhevsky2012imagenet}. The first DRL model to achieve the success of combing CNNs with RL is \cite{mnih2013playing}. Raw pixels in the game of Atari were fed into a CNN as input and the output of the CNN was a prediction of the future rewards.

Improving DRL with semi-supervised learning and self-supervised learning (SSL) is on the spotlight of the RL research field. For example, Deep Q-learning from Demonstrations (DQfD) uses small sets of demonstrations and accelerates the training through prioritized replay \cite{hester2018deep}. Self-Predictive Representations (SPR) is a recently proposed data-efficient method that relies on utilizing an encoder with the intentions to compute future latent state representations and to make predictions based on the learned transition model \cite{schwarzer2021dataefficient}.

We develop our novel DRL method combined with the concept of semi-supervised learning on the game of Othello, which is an unsolved board game. Board games, including Othello, were heavily experimented using DRL methods with game-specific settings \cite{8276588}. Our algorithm is strongly motivated by designing a computationally less costly RL method via using limited expert examples in a data-efficient manner. 

Semi-supervised learning uses a small amount of labeled data together with other unlabeled data, whereas SSL is a subset of unsupervised learning techniques. Semi-supervised learning focuses either on clustering or generating new labels from existing data. It consists of traditional methods like K-means \cite{wagstaff2001constrained} and more recent generative models like Generative Adversarial Networks (GANs) \cite{radford2015unsupervised}. K-means generates prototypes which are labeled based on the Euclidean metrics. GANs generate adversarial examples by utilizing both a generator and a discriminator. The generator attempts to generate fake examples that resemble the real ones, while the discriminator learns to distinguish the fake images from the real ones. As this process of adversary repeats, the generator eventually learns to generate realistic images.

The idea of performing control tasks by learning from expert examples was previously explored in Behavior Cloning. An agent learned a policy from expert behaviors in the form of state-action pairs \cite{pomerleau1991efficient}. An autonomous car was able to drive in a variety of road conditions with a speed of 20 miles per hour by behavior cloning from the samples of human drivers. As a drawback of supervised learning, the training of behavior cloning typically requires a large amount of data.

Moreover, semi-supervised learning can be an intuitive way to improve the performance of RL methods by the efficient usage of examples. Inspired by the ideas of combining semi-supervised learning with RL, several methods have been proposed. Inverse Reinforcement Learning (IRL) \cite{russell1998learning} makes agents adapt to uncertain environments and extract the reward signal (function) from straight observations of optimal expert behaviors \cite{ng2000algorithms}. The agent is able to choose actions based on the learned cost function. IRL has shown strength in many real-world applications. For example, IRL can successfully perform tasks like personalized route recommendation, traffic warning without any exact destinations, and battery usage optimization for hybrid electric vehicles, etc \cite{ziebart2008maximum}. IRL is a dual of an occupancy measure matching problem, and the learned cost function equals the dual optimum \cite{ho2016generative}.

However, IRL runs RL in the inner loop and hence has extremely high computational costs, and it lacks the ability to fit actions not present in the examples. The output of IRL is a cost function but does not make the agent choose an action in the environment. If the goal is to let an agent choose actions from the action space, a separate algorithm to let the agent take actions is needed. Generative Adversarial Imitation Learning (GAIL) \cite{ho2016generative} is an imitation learning algorithm inspired by the similarities between GANs and IRL. It takes the behaviors of the expert as input and gives the actions as output, skipping the part where the cost function is induced.

Q-learning is an asynchronous dynamic programming method that chooses actions based on the updated Q-values \cite{watkins1992q}. Q-learning uses an offline policy that is based on the Bellman Optimality Equation. In (\ref{eq:q}), $\gamma$ is the discounting factor, $Q^*(s,a)$ is the Q value of state $s$ with action $a$, $r_{t+1}$ is the result value at $s_{t+1}$.

\scriptsize
\begin{equation}
    Q^*(s,a) = E\{r_{t+1}+\gamma \mathrm{max}_{a'}Q^*(s_{t+1},a'|s_t=s,a_t=a)\}
    \label{eq:q}
\end{equation}
\normalsize

The method is described by \cite{mnih2013playing}. Deep Q-learning (DQN) is the NN implementation of Q-learning. For each episode of the game, the sampled data are not directly fed into the network, but kept in a replay buffer. A minibatch is then sampled from the replay buffer to train our NN. This improves both the sample efficiency and the sampling quality by reusing examples but decreasing the correlations between examples.

\section{Methods}
Our method is developed based on DQN \cite{mnih2013playing} and experimented using the game of Othello. Othello is an unsolved board game where there are two players playing against each other. An Othello board consists of a grid of $8\times8$ squares, with four squares in the middle set at the beginning of each game.

Whenever a piece is placed, all the opponent's pieces in between this piece and any of the player's pieces are flipped into the current player's color. A new piece can only be placed in a square that results in flips. If there is no valid move for one player, the turn is skipped. If there is no valid move for both players, the game ends. The final game result is determined by counting the pieces on the board. The player with the most pieces wins and a draw is also possible.

We use a variation of Double Q-learning (Double Q) \cite{hasselt2010double} and Dueling Q-learning (Dueling Q) \cite{wang2016dueling} to serve as our baseline model. We then design a novel algorithm which is suitable to incorporate expert examples. Afterwards, we train an agent either with or without expert examples and compare their performances against the baseline DQN.

For our algorithm, we tackle a plausible situation in which we have an offline expert, but the expert cannot tell the detailed Q-values (state-action values). Instead, it tells whether a state is a good state or not. This setting is especially realistic for many real world control tasks, where the reward function is hard to define and even the best expert cannot give a detailed evaluation on the state-action values. Moreover, it is often computationally less costly if an expert only provides a coarse evaluation of the state values instead of the accurate state-action values.


\subsection{Double and Dueling Q-learning}
Q-values in Q-learning will grow larger and larger since we only choose the maximal state-action value of the next state for each update, a.k.a. 'overestimation bias'. Overestimation bias often results in sub-optimal and unstable performance of Q-learning \cite{thrun1993issues}. The original Double Q exploits two different neural networks $Q^A$ and $Q^B$. In each update, only one network is chosen by some update scheme, e.g., random update. Either (\ref{eq:QA}) or (\ref{eq:QB}) will be used to update the chosen network by learning rate $\alpha(s,a)$, where $a^*$ is $\mathrm{argmax}_aQ_\theta^A(s',a)$, and $b^*$ is $\mathrm{argmax}_aQ_\theta^B(s',a)$ at next state $s'$.

\scriptsize
\begin{equation}
    \label{eq:QA}
    Q^A(s,a) = Q^A(s,a) + \alpha(s,a)(r+\gamma Q^B(s',a^*)-Q^A(s,a))
\end{equation}
\normalsize
\scriptsize
\begin{equation}
    \label{eq:QB}
    Q^B(s,a) = Q^B(s,a) + \alpha(s,a)(r+\gamma Q^A(s',b^*)-Q^B(s,a))
\end{equation}
\normalsize

Moreover, it is a common practice that the target network should just be a copy of the Q-network but with delayed synchronization, for the purpose of minimal computation overhead \cite{van2016deep}. Double Q introduces a new problem named underestimation, as it tends to underestimate Q-values. We use a variation of Double Q in our experiment to avoid underestimation, where we only update $Q_\theta^A$ and $a^*$ is $\mathrm{argmax}_aQ_\theta^B(s',a)$.

Dueling Q was specifically proposed based on DQN. The idea is that the state value $V(s)$ and the advantage $A^*(s,a)$ are implicitly contained in each Q-value $Q^*(s,a)$ by (\ref{eq:separate_value}). Hence, it is possible to represent those differently in an NN architecture in order to yield improved performance.
\scriptsize
\begin{equation}
    Q^*(s,a) =  A^*(s,a) + V(s)
    \label{eq:separate_value}
\end{equation}
\normalsize

\begin{figure}[t]
    \centering
    \includegraphics[width=0.8\linewidth]{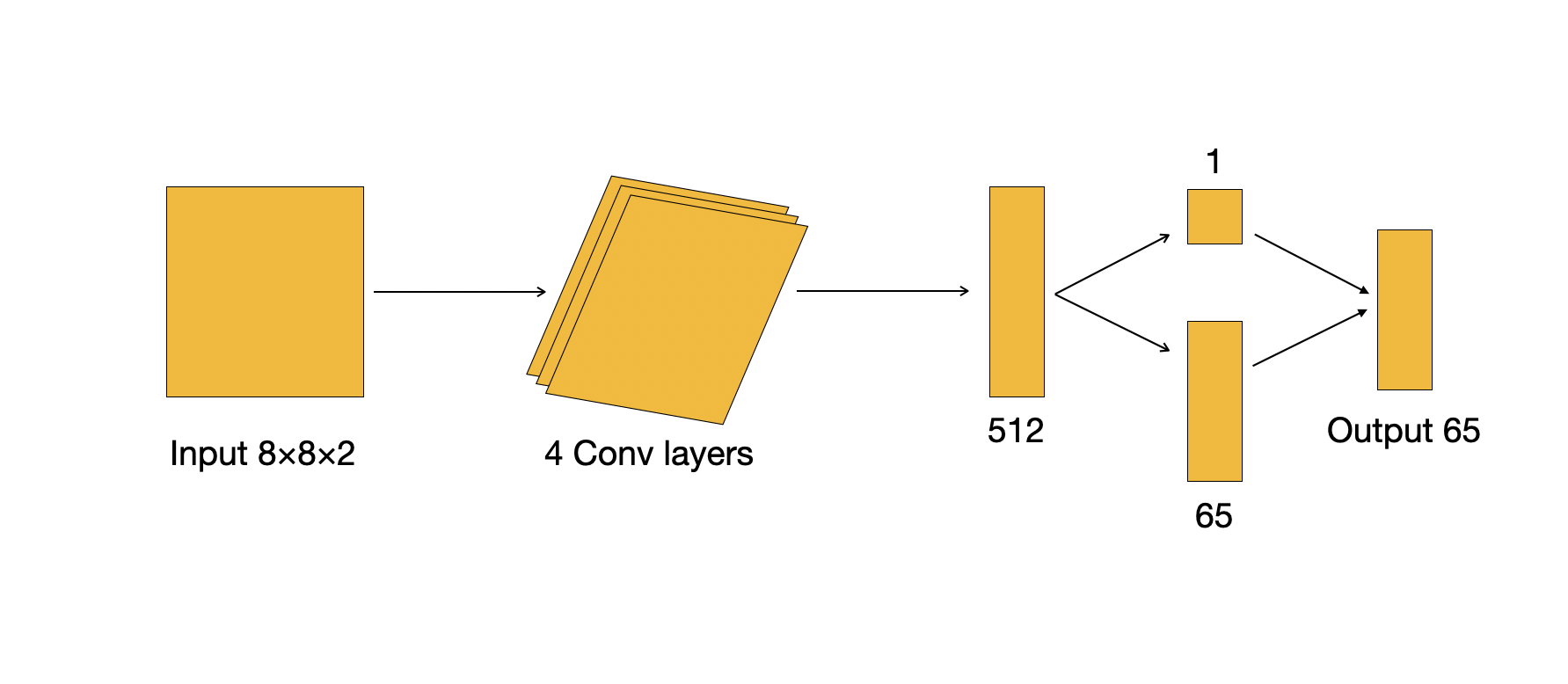}
    \caption{Architecture of our Dueling Q network. The fully connected layer splits into state value and action value branches, and their outputs are combined at the output layer in the Dueling Q network.}
\label{algo:dueling}
\end{figure}

The notion of maintaining two separate functions in Dueling Q traces back to \cite{baird1993advantage}, \cite{harmon1995advantage}, and \cite{harmon1996multi}. The shared Bellman residual update equation can be decomposed into one function with the state value and one function with the action advantage. Moreover, Dueling Q forces the maximal advantage to be 0 and replaces the $\mathrm{max}$ operator with an average when calculating the advantage, on purpose of improving the stability of algorithm. Compensating any change in the direction of the optimal action's advantage leads to potentially unstable training. Figure \ref{algo:dueling} illustrates our implementation of the Dueling Q architecture.

Consider the $i_{th}$ NN output $o_i$ in Dueling Q, $\omega_{i,a}$ the weight in the action layer, $\omega_{s}$ the state layer, and $K$ the total number of output neurons, then:
\scriptsize
\begin{equation}
o_i = \omega_{i,a}-\frac{1}{K}\sum_{k}^{K}\omega_{k,a} + \omega_{s}
\end{equation}
\normalsize

Suppose $\omega_{b}$ is the weight of the body layer ahead of branching, $y_i$ the target value of the $i_{th}$ neuron, then we have the gradient:
\scriptsize
\begin{equation}
\label{grad:dueling}
\begin{aligned}
    &\frac{\partial }{\partial \omega_{b}}(y_i-o_i)^2\\
    &=-2(y_i-o_i)(\frac{1}{K}\sum_{k}^{K}\frac{\partial o_i}{\partial \omega_{k,a}}\frac{\partial \omega_{k,a}}{\partial \omega_{b}}+\frac{\partial o_i}{\partial \omega_{s}}\frac{\partial \omega_{s}}{\partial \omega_{b}})\\
\end{aligned}
\end{equation}
\normalsize

In spite of the branching, the loss still back-propagates to the body layer as a whole during the procedure of gradient descent. This explains Dueling Q's inability in regulating overestimation bias of Q-learning. To improve on that, we decouple the gradient by using two different networks. The Q-values can be reduced when the gradients from two networks are independent because the overestimation bias would be represented by the state network instead.

\subsection{Expert Q-learning}

We propose our algorithm Expert Q-learning (Expert Q) to incorporate ideas from semi-supervised learning into Q-learning. The state values given by expert examples are from \{-1, 0, 1\}, simply indicating whether a state is good, neutral or bad. During training, we train an expert (state) network $E_\theta$ to predict the state values.

We improve the Dueling Q algorithm to utilize the state value in a more explicit way by composing directly the Q-values, so that the state value is not internally represented in the Q-network anymore, but appears in our update scheme. The equation of calculating our Q-value is shown in (\ref{eq:sate_value}). 

\scriptsize
\begin{equation}
    Q^*(s,a) =  Q_\theta(s,a) + E_\theta(s) - \frac{1}{K} \sum_{k}^{K}  Q_\theta(s,a_k)
    \label{eq:sate_value}
\end{equation}
\normalsize

Here, $K$ is the size of the action space, and $K=65$ since there is one action for each of the 64 squares on the board, plus an additional action of not being able to place a piece. Furthermore, $Q^*(s,a)$ is the updated Q-value by adding the state value $E_\theta(s)$ given by an expert, subtracting the mean of predictions at state $s$ from the network prediction $Q_\theta$.

Since we have $Q_\theta(s,a_k) = A(s,a_k) + V(s)$ by (\ref{eq:separate_value}), the predictions of the Q-network can indeed be treated the same as action advantages in Dueling Q during the update, by (\ref{eq:advantage_value}).
\scriptsize
\begin{equation}
\begin{aligned}
    &Q_\theta(s,a) - \frac{1}{K} \sum_{k}^{K}  Q_\theta(s,a_k)\\
    &=A(s,a) + V(s)-\frac{1}{K} \sum_{k}^{K} (A(s,a_k)+V(s))\\
    &=A(s,a) -\frac{1}{K} \sum_{k}^{K} A(s,a_k)
\end{aligned}
    \label{eq:advantage_value}
\end{equation}
\normalsize

\begin{algorithm}[tb]
\scriptsize
\caption{Expert Q-learning}\label{algo:mine}
\textbf{Input}: E, $L$, $\epsilon$, $\gamma$, $maxIter$\\
\textbf{Parameter}: D, $Q_\theta^A$,$Q_\theta^B$, $E_\theta^A$, $E_\theta^B$\\
\textbf{Output}: $Q_\theta^A$
\begin{algorithmic}[1]
\STATE $iter \gets 0$
\WHILE{$iter<maxIter$}
\WHILE{Game not finished}
\IF{$random()<\epsilon$}
\STATE Choose action randomly
\ELSE
\STATE Choose action based on $Q_\theta^A$
\ENDIF
\STATE Sample $s,a,s'$ into replay buffer D
\ENDWHILE
\STATE Add discounted $r$ for each state into replay buffer D
\STATE Sample minibatch of $s,a,s', r$ from D
\STATE $Q^A(s,a) \gets  Q_\theta^A(s,a) - \frac{1}{K} \sum_{k}^{K}  Q_\theta^A(s,a_k)$
\STATE $Q^A(s,a) \gets  Q^A(s,a) + E_\theta^A(s) $
\STATE $Q^B(s',a^*) \gets  Q_\theta^B(s',a^*) - \frac{1}{K} \sum_{k}^{K}  Q_\theta^B(s',a_k)$
\STATE $Q^B(s',a^*) \gets Q^B(s',a^*) + E_\theta^B(s')$
\STATE Update $Q_\theta^A, E_\theta^A$ by $L(r+\gamma Q^B(s',a^*), Q^A(s,a))$ with gradient descent
\IF {expert buffer E not empty}
\STATE Sample minibatch of $s^e,v^e$ from E
\STATE Update $E_\theta^A$ by $L(E_\theta^B(s^e), v^e)$ with gradient descent
\ENDIF
\IF {$Synchronize$}
\STATE $Q_\theta^B \gets Q_\theta^A$, $E_\theta^B \gets E_\theta^A$
\ENDIF
\STATE $iter++$
\ENDWHILE
\STATE \textbf{return} $Q_\theta^A$
\end{algorithmic}
\end{algorithm}
\normalsize

Our full method is described in Algorithm \ref{algo:mine}. The first part of the algorithm, i.e., line 2 to line 11 is similar to the steps of Double Q, while our Expert Q steps follow from line 12 to line 21. In the algorithm, $r$ is $\gamma$ the discounted result value, $\gamma$ is the discounting factor, $\epsilon$ is the exploration ratio, $s'$ is the next state, $a^*$ is $\mathrm{argmax}_aQ_\theta^B(s',a)$, $maxIter$ is the maximal iterations, and $L()$ is the loss function, i.e., Mean Squared Error (MSE) in this case.

We have four networks  $Q_\theta^A$,$Q_\theta^B$, $E_\theta^A$ and $E_\theta^B$, where $Q_\theta^A$ and $E_\theta^A$ are the Q-network and expert (state) network, $Q_\theta^B$ and $E_\theta^B$ are their copies. $E_\theta^A$ is trained with sate $s^e$ and state value $v^e$ when the expert example buffer $E$ is not empty.

As mentioned above, our overestimation bias is represent in the expert network instead. The update in line 20 of Algorithm \ref{algo:mine} has the effect of reducing overestimation bias because discrete values from \{-1, 0, 1\} are used as the target values of the expert network, which would be considerably smaller than the actual output when the overestimation happens.

It is possible to implement our algorithm without using expert examples in line 18 to line 21 of Algorithm \ref{algo:mine}, which is named Expert Q without examples. The effects of not using expert examples are deeply explored in our experiment because we would like to perform an ablation study about the usage of expert examples.

\subsection{Experimental Setup}
In our experiment, positions of board games are represented by binary values in two channels, with each channel denoting the pieces of one player, which amounts to an input size of $8\times8\times2$. The outputs are values corresponding to actions, which amounts to an output size of $8\times8+1$.

While DQN models represent the players, we consider as opponent a random player (RANDOM), a greedy player (GREEDY), and a stochastic player (STOCHASTIC). GREEDY chooses an action simply by checking one move ahead. STOCHASTIC plays 70\% randomly as RANDOM does, and 30\% greedily with the weight of each cell determined according to the heuristic 'HEUR' described in \cite{van2013reinforcement}. The heuristic value of each cell on the Othello board is shown in Table \ref{table:heur}.

Table \ref{table:heur} also plays the role of an 'expert' in our experiment, giving out the state value from \{-1, 0, 1\}. When playing against RANDOM, expert examples are sampled from STOCHASTIC playing against RANDOM. The examples are sampled from STOCHASTIC playing against GREEDY when the opponent is GREEDY. They are sampled from STOCHASTIC playing against STOCHASTIC when the opponent is STOCHASTIC.

It was found that Q-learning performed the best when it was trained and tested against a fixed player in \cite{van2013reinforcement}. Thus, we train and test a model by playing against the same opponent.

\begin{table}[t]
\scriptsize
\centering
\begin{tabular}{ |c|c|c|c|c|c|c|c|} 
 \hline
100 & -25 & 10 & 5& 5& 10& -25& 100\\
 \hline
 -25& -25& 2& 2& 2& 2& -25& -25\\
  \hline
10& 2& 5& 1& 1& 5& 2& 10\\
 \hline
5& 2& 1& 2& 2& 1& 2& 5\\
 \hline
5& 2& 1& 2& 2& 1& 2& 5\\
 \hline
10& 2& 5& 1& 1& 5& 2& 10\\
 \hline
-25& -25& 2& 2& 2& 2& -25& -25\\
 \hline
100& -25& 10& 5& 5& 10& -25& 100\\
 \hline
\end{tabular}
\caption{Heuristic table for Othello. The value in each cell represents the weight of the piece in this square.}
\label{table:heur}
\end{table}

Players always play from the initial board position during training, but they play from all 236 initializations of positions during testing. There are unique 236 board positions after 4 turns from the start of Othello \cite{van2013reinforcement}. Testing all 236 initializations yields results that can reflect the capabilities of models more properly. Each player plays the white and black side evenly in one round during testing, which is in total a number of 472 games per round. Results are measured by a score calculated from the number of wins and draws out of total number of games, as shown by (\ref{eq:score}):

\scriptsize
\begin{equation}
\label{eq:score}
    score = \frac{wins+0.5\times draws}{games}
\end{equation}
\normalsize

\subsection{Parameters and Architecture}
Most parameters are kept as the same as possible across experiments. The discounting factor $\gamma$ is set to 0.99. In the meantime, the reward at a single state is also the $\gamma$ discounted value of the final outcome of that game. The learning rate is chosen as $1e^{-4}$ for both the state network and Q-network. The total number of iterations is $100\times10^3$ and the copies of networks are synchronized every $2\times10^3$ iterations, where the exploration ratio $\epsilon$ decreases linearly from 1 to 0.01 during training. Both the buffer sizes of replay memories and expert examples are set to 10000, with a minibatch size of 64. Expert examples are randomly selected 10000 out of 100000 samples to maintain the intention of using a replay buffer, i.e., decreasing correlations between examples.

\begin{figure}[t]
    \centering
    \includegraphics[width=0.8\linewidth]{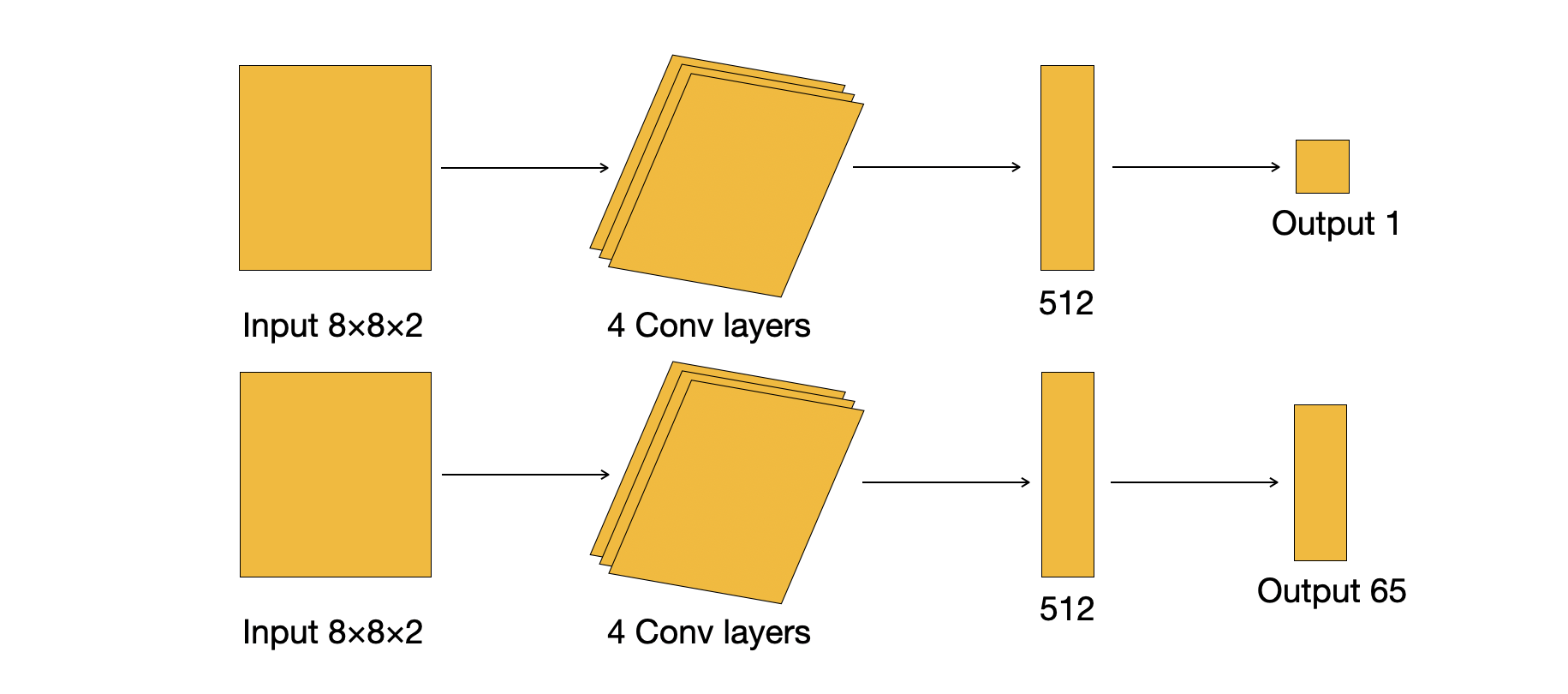}
    \caption{Architecture of our Expert Q. Convolutional layers have 64 filters with kernel size (3,3) and stride 1. The first convolutional layer uses a padding of 1. A fully connected layer with 512 output, dropout rate of 0.3 and a sigmoid activation function immediately follows. Those layers are combined with batch normalization.}
\label{algo:expertfig}
\end{figure}

Our network consists of four 2D convolutional layers. The state network has an additional output layer with output size 1 and the Q-network has an additional layer of output size 65. Double Dueling Q uses an action branch with output size 65 and a state branch with output size 1. The final output layer is also with output size 65. The architecture of Expert Q is shown in Figure \ref{algo:expertfig}. The top is the expert network and the bottom is the Q-network. Most of the layers are kept as the same as possible.

\section{Results}
 
\begin{figure*}[t]
    \centering
    \includegraphics[width=0.8\linewidth]{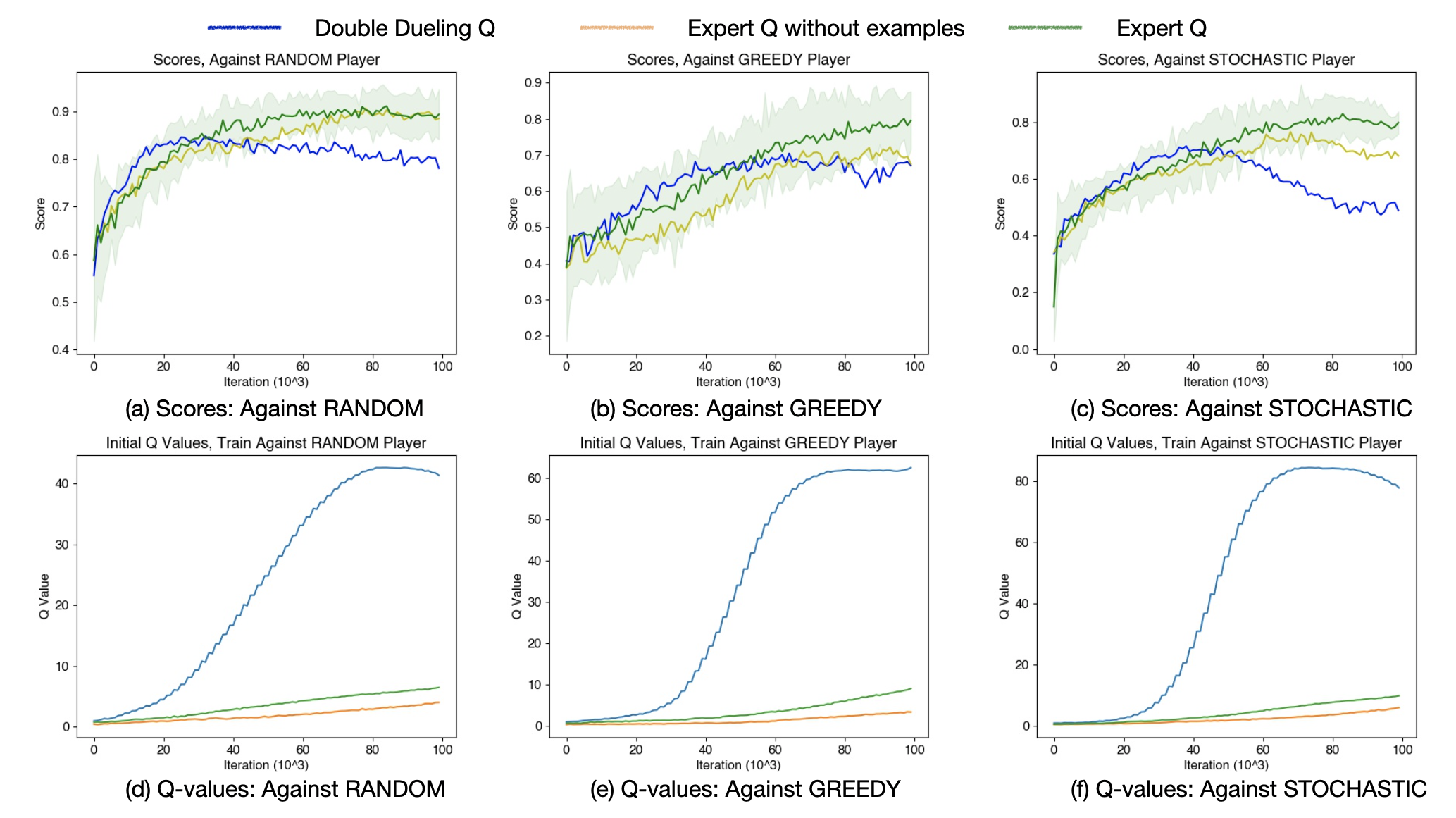}
    \caption{Results when playing against different opponents. The top ones are the scores and the bottom ones are initial Q-values, which are the predictions of Q-values of the initial board position by the Q-network. Only the 95\% confidence interval of Expert Q scores is plotted for the sake of clarity.}
\label{fig:result1}
\end{figure*}

The results are obtained from playing against RANDOM, GREEDY and STOCHASTIC. In each setting, the plot is an average of 10 different results, shown in Figure \ref{fig:result1}. Means ($\mu$) and standard deviations ($\sigma$) after iterations of $50\times10^3$ and $100\times10^3$ are shown in Table \ref{table:results}.

Results illustrate that Expert Q performs the best out of three models after $100\times10^3$ iterations of training when it is trained and tested against the same opponent in all three experiment settings. Expert Q without examples performs overall slightly worse than Expert Q, but still shows better performance than Double Dueling Q except when playing against GREEDY.

Double Dueling Q has a sudden drop of scores after approximately $40\times10^3$ iterations in Figure \ref{fig:result1} (c) whereas its score stays more reasonable in Figure \ref{fig:result1} (b). In the meantime, the initial Q-values of Double Dueling Q in Figure \ref{fig:result1} (d), (e) and (f) rise drastically after $40\times10^3$ iterations. The scores of Double Dueling Q also rise faster than the other two models in the beginning of all three opponent settings. This can be attributed to its quickly rising Q-values, which is an indicator of both faster convergence and higher overestimation bias.

Figure \ref{fig:result1} (d), (e) and (f) all show that the Q-values in Expert Q are much smaller than in Double Dueling Q. Expert Q has slightly larger initial Q-values than Expert Q without examples because its state network is trained on expert examples, which introduces slightly more bias into the Q-network.

The three different settings of the opponent (RANDOM, GREEDY, STOCHASTIC) correspond to a random environment, a deterministic environment and a stochastic environment, respectively. The Double Dueling Q model makes more mistakes in the stochastic environment than in the deterministic environment when the choices are based on overoptimistic Q-values. This result corresponds to the previous conclusion that the issue of overestimation bias was found to be most salient in the stochastic environment \cite{chen2021investigation}.

\begin{table*}[t]
\scriptsize
\centering
\begin{tabular}{ |p{4cm}||p{1cm}|p{1cm}|p{1cm}|p{1cm}|p{1cm}|p{1cm}|} 
 \hline
 &
 \multicolumn{2}{|c|}{RANDOM} &
 \multicolumn{2}{|c|}{GREEDY} &
 \multicolumn{2}{|c|}{STOCHASTIC} \\
 \hline
 Iteration ($10^3$)&50&100&50&100&50&100\\
 \hline
 Double Dueling Q& $0.83\pm 0.03$& $0.78\pm 0.07$& $0.66\pm 0.05$& $0.67\pm 0.11$& $0.68\pm 0.06$ &$0.49\pm 0.05$\\
 \hline
 Expert Q without examples& $0.85\pm 0.02$& $0.88\pm 0.02$& $0.61\pm 0.05$& $0.68\pm 0.04$& $0.68\pm 0.05$ &$0.68\pm 0.06$\\
 \hline
 Expert Q& $0.87\pm 0.03$& $0.89\pm 0.03$& $0.66\pm 0.06$& $0.8\pm 0.04$& $0.73\pm 0.07$ &$0.8\pm 0.02$\\
 \hline
\end{tabular}
\caption{Scores with means and standard deviations ($\mu\pm \sigma$) calculated.}
\label{table:results}
\end{table*}

It can be informative to check the results of trained models playing against each other. Table \ref{pit:stochastic} shows the results of players trained against STOCHASTIC playing against each other in 10 rounds, a total of 4720 games. We find out that the baseline player (Double Dueling Q) actually performs almost equally to Expert Q without examples when the opponent is Expert Q, despite the fact that Double Dueling Q has poor performance when playing against Expert Q without examples. In turn, the Expert Q player demonstrates the best performance as expected.

The result can be an illustration of bias-variance trade-off. High bias in our baseline model leads not only to poor performance when trained and tested against STOCHASTIC, but also to relatively more stable performance when playing against other players. This corresponds to the observations that the improvement of performance is more significant when expert examples are used to train the state network.

\begin{table*}[t]
\scriptsize
\centering
\begin{tabular}{ |c||c|c|c|} 
 \hline
   & Double Dueling Q & Expert Q without examples & Expert Q  \\ 
  \hline
 Double Dueling Q &  & 1080 (0.23) & 917 (0.19) \\ 
  \hline
 Expert Q without examples & 3556 (0.75)&  &890 (0.19)\\ 
 \hline
 Expert Q & 3753 (0.8)&3750 (0.79)& \\ 
 \hline
\end{tabular}
\caption{Players play against each other in 4720 games. Each entry represents the wins and winning percentage of the player in that row against the player in that column. Draws are not included in the table.}
\label{pit:stochastic}
\end{table*}

\section{Discussion}

Our results indicate that Expert Q holds significant advantages over Double Dueling Q especially when playing against a stochastic player (a stochastic environment). Expert Q without examples demonstrates superior performance than Double Dueling Q when it is trained and tested against a non-deterministic player.

When our trained models play against each other, Expert Q still maintains the highest performance out of three models, whereas Expert Q without examples wins Double Dueling Q but does not show more strength when they play against Expert Q. This can be explained by the bias-variance trade-off and shows exactly that Expert Q is an algorithm designed in joint use with expert examples.

In Q-learning, high bias is unavoidable and often not desired. Our experiment, however, demonstrates that higher bias during training or even testing results in more stable performance if the bias is from expert examples. The performance of a model (Expert Q) with slightly higher bias than Expert Q without examples provides the best performance out of three models. We argue that low bias is not always desirable, especially when the environment is susceptible to change, and somewhat higher bias can make the model more resistant to those changes.

It is worth mentioning that Double Q can be underestimating instead of overestimating when $Q_\theta^A$ is used to chose $a^*$. The performance also depends on the choice of parameters (e.g., learning rate, synchronization scheme).

To further incorporate SSL into our method, it might be interesting to adapt the Teacher-Student paradigm into our expert network, where a teacher is trained by existing expert examples, and then used to label unlabeled new examples \cite{yalniz2019billion}. Afterwards, a student will be trained from those data labeled by the teacher. Meanwhile, the student can be designed with increased noise during the process of training \cite{xie2020self} to guarantee the learning of the real situation.

\section{Conclusion}
Our research contributes a novel algorithm that exploits state values from expert examples and uses relatively fewer computational resources than MCTS in decision making games. Our trained models are put directly into contests on Othello, which otherwise cannot be achieved if the experiment was performed in a typical single-player game environment. 

Expert Q shows stronger performance comparing to the baseline algorithm and bests both Doubling Dueling Q and Expert Q without examples in direct competitions. The use of expert examples improves the performance by a large margin compared to not using expert examples.

The combined use of semi-supervised learning and DRL is expected to be applicable in many real world control tasks, such as autonomous driving and unmanned exploration.

\section{Acknowledgments}
The research presented in this paper has benefited from the Experimental Infrastructure for Exploration of Exascale Computing (eX3), which is financially supported by the Research Council of Norway under contract 270053. We want to acknowledge the help received from the Department for Research Computing at USIT,
the University of Oslo IT-department. This work was performed on the [ML node] resource, owned by the University of Oslo, and operated by the Department for Research Computing at USIT,
the University of Oslo IT-department. http://www.hpc.uio.no/

\bibliographystyle{abbrvnat}
\bibliography{main}

\end{document}